\documentclass[letterpaper, 10 pt, conference]{ieeeconf}  % Comment this line out if you need a4paper

\IEEEoverridecommandlockouts                              % This command is only needed if 
\usepackage{graphicx} % for pdf, bitmapped graphics files
\usepackage{amsmath} % assumes amsmath package installed
\usepackage{amssymb}  % assumes amsmath package installed

\title{\LARGE \bf
An Interactive Hands-Free Controller for a Riding Ballbot to Enable Simple Shared Control Tasks
}

\author{Chenzhang Xiao$^{1*}$, Seung Yun Song$^{1}$, Yu Chen$^{1}$, Mahshid Mansouri$^{1}$, \\Jo\~{a}o Ramos$^{1}$, William R. Norris$^{2}$, and Elizabeth T. Hsiao-Wecksler$^{1}$% <-this % stops a space
\thanks{This work was supported by NSF Grant No. 2024905.}% <-this % stops a space
\thanks{$^{1}$Mechanical Science and Engineering, University of Illinois at Urbana-Champaign, Urbana, IL 61801 USA email: (cxiao3, ssong47, yuc6, mm64, jlramos, ethw)@illinois.edu. *Corresponding author.}
\thanks{$^{2}$Industrial and Enterprise Systems Engineering, UIUC, Champaign, IL 61820 USA (wrnorris@illinois.edu)}}%

\begin{document}

\maketitle
\thispagestyle{empty}
\pagestyle{empty}

%%%%%%%%%%%%%%%%%%%%%%%%%%%%%%%%%%%%%%%%%%%%%%%%%%%%%%%%%%%%%%%%%%%%%%%%%%%%%%%%
\begin{abstract}

Our team developed a riding ballbot (called PURE) that is dynamically stable, omnidirectional, and driven by lean-to-steer control. A hands-free admittance control scheme (HACS) was previously integrated to allow riders with different torso functions to control the robot's movements via torso leaning and twisting. Such an interface requires motor coordination skills and could result in collisions with obstacles due to low proficiency. Hence, a shared controller (SC) that limits the speed of PURE could be helpful to ensure the safety of riders. However, the self-balancing dynamics of PURE could result in a weak control authority of its motion, in which the torso motion of the rider could easily result in poor tracking of the command speed dictated by the shared controller. Thus, we proposed an interactive hands-free admittance control scheme (iHACS), which added two modules to the HACS to improve the speed-tracking performance of PURE: control gain personalization module and interaction compensation module. Human riding tests of simple tasks, idle-keeping and speed-limiting, were conducted to compare the performance of HACS and iHACS. Two manual wheelchair users and two able-bodied individuals participated in this study. They were instructed to use ``adversarial" torso motions that would tax the SC's ability to keep the ballbot idling or below a set speed, i.e., competing objectives between rider and robot. In the idle-keeping tasks, iHACS demonstrated minimal translational motion and low command speed tracking RMSE, even with significant torso lean angles. During the speed-limiting task, where the commanded speed was saturated at 0.5 m/s, the system achieved an average maximum speed of 1.1 m/s with iHACS, compared with that of over 1.9 m/s with HACS. These results suggest that iHACS can enhance PURE's control authority over the rider, which enables PURE to provide physical interactions back to the rider and results in a collaborative rider-robot synergy.
\end{abstract}

%%%%%%%%%%%%%%%%%%%%%%%%%%%%%%%%%%%%%%%%%%%%%%%%%%%%%%%%%%%%%%%%%%%%%%%%%%%%%%%%
\section{INTRODUCTION}

\begin{figure}[thpb]
  \centering
  \includegraphics[scale=0.55]{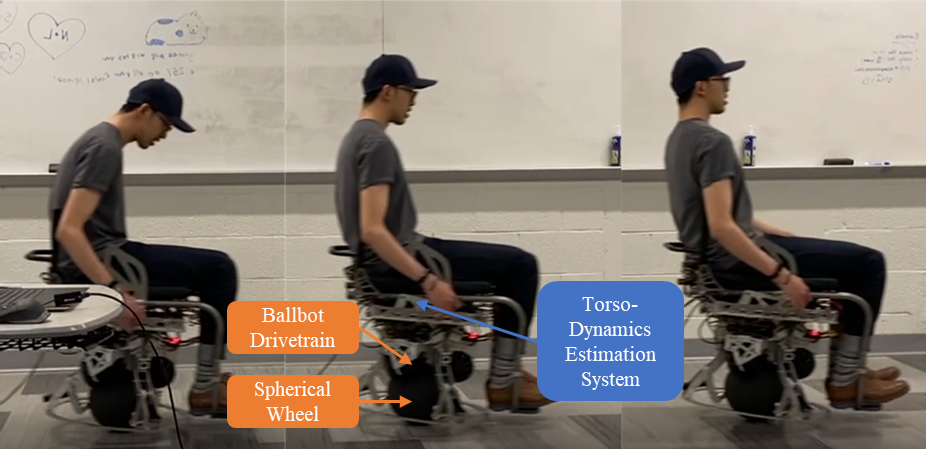}
  \caption{Rider on PURE during acceleration-deceleration test translating forward. The movement is controlled via torso learning, i.e., forward to drive, backward to brake. PURE is composed of a ballbot drivetrain that uses a spherical wheel for maneuvering, and a torso-dynamics estimation system (TES) to achieve a hands-free lean control with tunable sensitivity.}
  \label{overview}
\end{figure}

Our group has developed a riding ballbot called PURE, personal unique rolling experience, where riders control the device's omnidirectional movement through torso leaning (for translation) and twisting (for spinning)  (Fig. \ref{overview}) \cite{xiao2022personal}. 
The hardware platform features a custom ballbot drivetrain, i.e., a dynamically-stable mobile robot that balances on a ball as its driving wheel, with a similar design as described in \cite{xiao2023design}. A seating module known as the Torso-dynamics Estimation System (TES) was also integrated to measure rider torso motion and physical human-robot interactions \cite{song2023design}. A low-level balancing controller, incorporating a cascaded LQR-PI loop, was developed for the balancing control of the ballbot drivetrain, achieving a top speed of 2.4 m/s and a braking time of 2 seconds from 1.4 m/s while carrying a 60 kg payload \cite{xiao2022personal}. Additionally, a mid-level interface controller, the Hands-Free Admittance Control Scheme (HACS), was implemented to use physical human-robot interactions (pHRI) measured by the TES for command speed generation, allowing control of the ballbot through torso movements \cite{song2024driving}. This control scheme provided adjustable system responses to accommodate riders with varying levels of torso motion capabilities. Both able-bodied individuals and wheelchair users have successfully navigated indoor tasks of varying difficulties riding PURE \cite{song2024driving}. However, the system still demands a high degree of motor coordination and proficiency from the rider for precise control of PURE’s movement. During navigation studies, some “collision events” were recorded in constrained spaces. To address this, we envisioned a shared controller (SC) could enhance safety and reduce rider proficiency requirements.

\begin{figure}[thpb]
  \centering
  \includegraphics[scale=0.7]{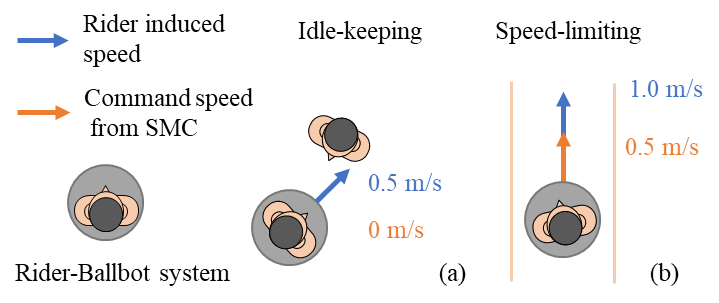}
  \caption{Simple illustrations of possible shared controller (SC) to enhance rider safety. (a) Idling scene: SC sets the command speed to zero when the system aims to remain in a fixed location, resisting any translational motion induced by rider torso movement. (b) Speed-limiting scene: SC adjusts the magnitude of the translational movement to avoid collision.}
  \label{SC_concept}
\end{figure}

Shared control (SC) is a generic concept in which multiple entities can coordinate their movements in a shared environment \cite{marcano2020review}. 
In the context of the automotive industry, one embodiment of SC is the advanced driver assistance system, which has been widely used to provide safety for passenger vehicles. It allows the vehicle to temporarily share the control authority with the driver for safeguarding such as lane keeping and emergency braking \cite{driver_assis_sys}.
% Provide more literature review on shared-control
%
For PURE, we envisioned SC could be applied to PURE riding as a motion planner to regulate command speed generated from HACS. For example, when the user is riding PURE and idling in front of an object, the SC should keep a zero command speed in the direction that could cause a potential collision, regardless of any candidate command speed generated from rider's torso motion (Fig. \ref{SC_concept}). When the rider is navigating PURE in a narrow hallway, the SC should limit the ballbot's top speed to minimize the risk of collision (Fig. \ref{SC_concept}b)

Such shared control concepts have been explored in research communities focusing on four-wheel statically-stable wheelchairs, to assist with obstacle avoidance \cite{trieu2008shared, herrera2018modeling}, swap control from rider to wheelchair in emergency situations \cite{urdiales2011new}, and reduce control uncertainty when the wheelchair is controlled with novel brain-machine interface \cite{philips2007adaptive, li2016human} or voice command \cite{simpson1997adaptive}. SC demonstrated promising results in enhancing system safety and reducing rider proficiency requirements for navigation; however, these methods are not directly applicable to PURE due to its unique dynamic stability as a ballbot.

Indeed, the effectiveness of SC relies not only on its perception of and strategic maneuvering through the surrounding environment, but also on the drivetrain’s capacity to accurately follow these modulated commands. However, command speed tracking has been problematic for underactuated systems when external forces or moments are present \cite{underactuated}. 
For example, due to the dynamics of the system, the self-balancing controller of a ballbot induces a translational motion when an external load is applied to change the drivetrain tilt angle, even though the command signals for chassis tilt and translational speed are kept at zero.
Such behavior was exploited to provide hands-free lean-to-steer control of its translational motion, but could also hinder the effectiveness of any SC due to diminished capability to track a command translational speed. In our prior studies, simulation experiments of PURE found that the rider could drive PURE to 1.4 m/s by torso leaning while the command speed was set to zero \cite{xiao2022personal}. Experiments with the physical hardware of PURE also revealed that the measured speed consistently exceeded the command speed \cite{xiao2022personal}.
These observations highlight the limitations of the current Hands-Free Admittance Control Scheme (HACS) with LQR-PI controller, which does not adequately account for the dynamics of torso leaning. Consequently, maintaining accurate speed tracking with the rider-ballbot system remains challenging.

The drivetrain’s poor tracking performance can be attributed to inaccuracies in the system model used by the LQR-PI controller and, more significantly, to the applied forces and torques exerted on PURE by the rider. To enhance the model accuracy, biomechanical models can be employed to estimate the rider-ballbot system’s mass, center of mass location, and inertia for each individual rider. To address the impact of physical interaction forces and torques, we propose evaluating PURE’s system dynamics under these forces and determining new equilibrium conditions necessary to maintain dynamic stability. Thus, this study introduces a new control scheme, the Interactive Hands-Free Admittance Control Scheme (iHACS), which incorporates these considerations. To assess the effectiveness of iHACS, we conducted human subject tests involving simple SC tasks: idle-keeping and speed-limiting.

The rest of this paper is structured as follows: Section II presents the modeling of the ballbot, incorporating externally applied interaction forces and torques. Section III introduces the proposed iHACS. Section IV details the human subject tests, in which test participants rode PURE using the HACS and iHACS schemes during simple SC tasks, followed by discussions in Section V and conclusions in Section VI.

\section{MODELING OF BALLBOT WITH EXTERNAL INTERACTIONS}

\begin{figure}[thpb]
  \centering
  \includegraphics[scale=0.5]{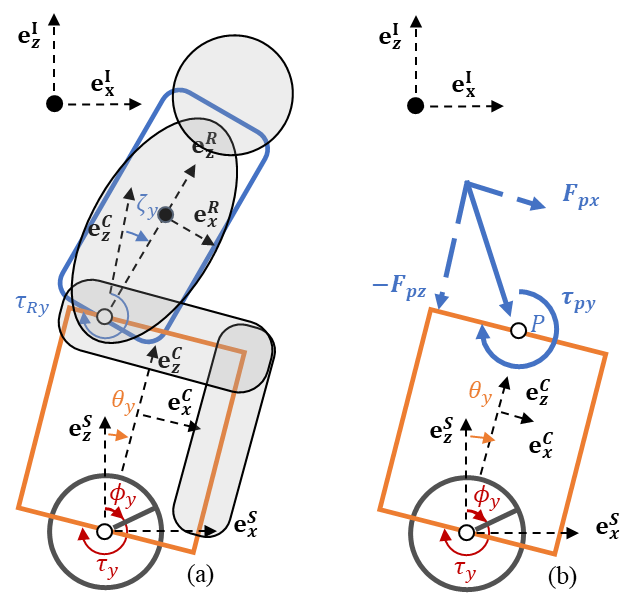}
  \caption{Modeling of rider-ballbot system in the sagittal plane. (a) Complete model of rider and PURE. (b) Isolated model of PURE: interaction between robot and rider represented by forces and moments applied at point P.}
  \label{model}
\end{figure}

We investigated a dynamic model for rider-ballbot system in the sagittal plane, considering the effects of rider-ballbot interaction forces and torques applied to the chassis drivetrain via torso leaning. For this evaluation, we isolated the ballbot from the complete rider-ballbot model (Fig. \ref{model}a) and assumed all interaction forces and torques from rider to the ballbot were applied at the top center of the chassis, denoted as point P (Fig. \ref{model}b).

Using Euler-Lagrange's method \cite{spong2004robot}, the equations of motion for the planar ballbot model were derived by incorporating externally applied physical interaction forces and torques. This model had two generalized coordinates, $\mathbf{q_y} = [\theta_y, \phi_y]^T$, where $\theta_y$ is chassis tilt angle and $\phi_y$ is the spherical wheel (SW) angular displacement (Fig. \ref{model}).
Inputs to the system ($\mathbf{u_{Iy}}$) include the input torque applied to the SW ($\tau_y$) and physical human-robot interactions $\mathbf{u_{py}}$ applied to the chassis by the rider as defined in the chassis coordinate frame (Fig. \ref{model}b), which contains 2D forces ($F_{px}$, $F_{pz}$) and torque ($\tau_{py}$). Hence,
\begin{equation}
    \mathbf{u_{Iy}} = [\tau_y, \mathbf{u_{py}}^T]^T = [\tau_y, F_{px}, F_{pz}, \tau_{py}]^T
\end{equation}
In this case, we can solve for the equation of motion of the system in the sagittal plane to be:
\begin{equation}
    \mathbf{M_{Iy}(q_y,\dot{q}_y)\ddot{q}_y + C_{Iy}(q_y,\dot{q}_y) + G_{Iy}(q_y) = J_{Iy}^T u_{Iy}}
    \label{eqn3:EOM}
\end{equation}
where $\mathbf{M_{Iy}}$ is the mass and inertia matrix, $\mathbf{C_{Iy}}$ is the Coriolis matrix, $\mathbf{G_{Iy}}$ is the gravity matrix, and $\mathbf{J_{Iy}}$ is the Jacobian matrix. Thus, the system dynamics equation of the ballbot drivetrain subject to physical interactions ($f_{Iy}$) can be linearized and derived as
\begin{equation}
    \mathbf{\ddot{q}_{y}} = f_{Iy}(\mathbf{q_y,\dot{q}_y,u_{Iy}})
    \label{eqn3:dynamics}
\end{equation}

To achieve dynamic stability in the sagittal plane, we further investigated the equilibrium conditions using the derived system dynamics. Our focus was on a dynamic equilibrium where the ballbot maintains a constant tilt angle ($\ddot\theta_y = \dot\theta_y = 0$) for a given command SW angular speed ($\ddot\phi_y = 0, \dot\phi_y =\dot\phi_{cy}$). This required providing a constant input SW torque ($\tau_{EQy}$) that compensated for the externally applied physical interaction forces and torques from the rider. 
Hence, (\ref{eqn3:dynamics}) becomes
\begin{equation}
\begin{aligned}
    & [0,0]^T = f_{Iy}(\mathbf{q_y,\dot{q}_y,u_{Iy}}) \\
    & = f_{Iy}([\theta_{{EQy}}, \phi_{EQy}]^T,[0, \dot\phi_{cy}]^T, [\tau_{{EQy}}, \mathbf{u_{py}}^T]^T)
\end{aligned}
\label{eqn3:equilibirium}
\end{equation}
Note the angular displacement of SW when maintaining such dynamic equilibrium ($\phi_{EQy}$) can be arbitrary values since this term does not appear in the (\ref{eqn3:equilibirium}) when fully expanded. We used $\phi_{EQy} = 0$ in this study.

This equation offers insights into the dynamics of the ballbot in the presence of external interaction forces and torques. To maintain a constant translational speed $\dot\phi_{cy}$ (or be stationary when $\dot\phi_{cy}=0$), the ballbot requires a specific tilt angle $\theta_{{EQ_y}}$ and input torque $\tau_{{EQ_y}}$ to counteract the externally applied interactions $\mathbf{u_{py}}$. (\ref{eqn3:equilibirium}) is re-arranged to pull $\theta_{{EQ_y}}$ and $\tau_{{EQ_y}}$ to the left side of the equation. 
\begin{equation}
    [\theta_{{EQy}}, \tau_{{EQy}}]^T = f_{EQy}(\mathbf{u_{py}, \dot\phi_{cy}})
    \label{eqn3:inter_comp}
\end{equation}
In the next section, this concept forms the basis of our interactive hands-free admittance controller.

\section{CONTROL SYSTEM DEVELOPMENT}

The proposed interactive Hands-Free Admittance Control Scheme (iHACS) builds on the Hands-Free Admittance Control Scheme (HACS) (Fig. \ref{block_diagram}a). iHACS incorporates two additional modules: an offline control gain personalization module, which customizes control gains for each rider, and an online interaction-compensation module.

\subsection{Control Gain Personalization}

The control gain personalization module allows adjustments to the LQR control gains based on the rider's weight and height. In a prior study \cite{xiao2022personal}, this method was used to derive control gains for a rider weighing 60 kg and measuring 1.8 m, which were then applied to the LQR gains in the baseline controller for all participants in the experimental tests. In this study, we explicitly derived the control gains for the individual participant based on the self-reported height and measured weight (averaged from TES measurements when in park mode). Thus, the rider's inertia was calculated from anatomic data presented in \cite{yoganandan2009physical} and lumped with the mass and inertia of ballbot drivetrain. Lastly, the sagittal plane equation of motion for the rider-ballbot system was derived using a wheeled-inverted-pendulum model \cite{xiao2023design}, with state-space vector $\mathbf{s_y}=[\theta_y, \phi_y, \dot\theta_y, \dot\phi_y]^T$. The optimal control gains for LQR ($\mathbf{k_{LQR}^*}$) were obtained by solving the Riccati differential equation \cite{liberzon2011calculus}. Note that we do not control the angular displacement of SW ($\phi_y$) directly, hence the quadratic cost, control gain, and command signal associated with $\phi_y$ are all zero for this study.

\subsection{Interaction Compensation}

In the interaction compensation module, we used (\ref{eqn3:inter_comp}) to calculate a new equilibrium tilt angle $\theta_{EQy}$ and input torque $\tau_{EQy}$ based on measured interaction forces and torques vector $\mathbf{u_{py}}$ and command speed $\dot\phi_{cy}$. 
That is, in order to track a command speed of $\dot\phi_{cy}$ with the presence of applied interaction forces and torques $\mathbf{u_{py}}$, the PURE chassis requires a tilting angle of $\theta_{EQy}$ and an additional input torque of $\tau_{EQy}$.
Assuming that the rider's movement is at a much lower dynamic bandwidth compared with the PURE drivetrain, we proposed to directly modify the command state signal for the LQR controller from $\mathbf{s_{cy}} = [0, 0, 0, \dot\phi_{cy}]$ for HACS to  
\begin{equation}
        \mathbf{s_{cy}^*} = [\theta_{EQy}, 0, 0, \dot\phi_{cy}] \\
\end{equation}

In addition, $\tau_{EQy}$ will also be added to the reference input torque calculated from the LQR controller.

\subsection{Control System Integration}

\begin{figure}[thpb]
  \centering
  \includegraphics[scale=0.55]{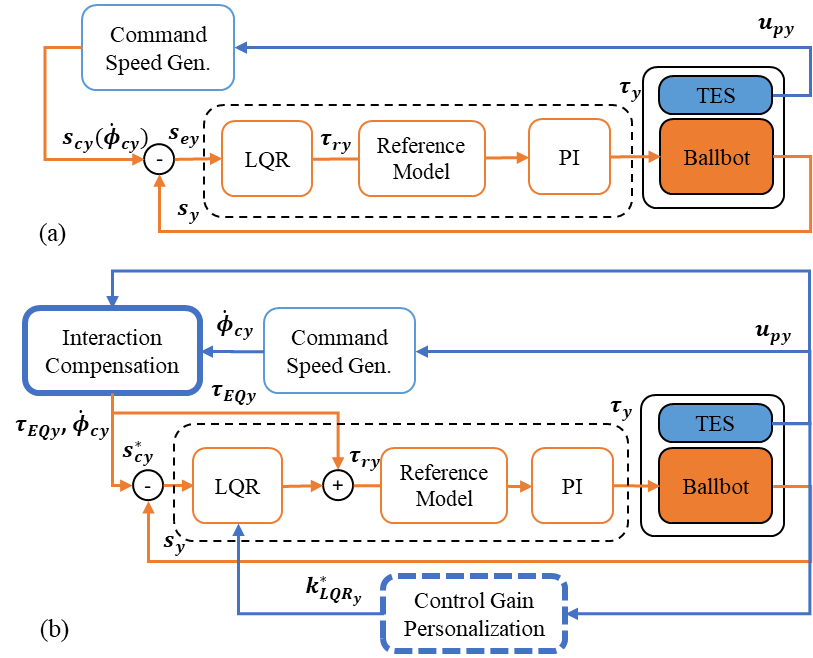}
  \caption{Control block diagram for (a) hand-free admittance control scheme (HACS) and (b) the proposed interactive hands-free admittance control scheme (iHACS). The command speed generator has tunable parameters and calculates the command translational speed for the low-level LQR-PI baseline balance controller for the ballbot drivetrain. The iHACS includes two additional modules: (1) the control gain personalization module that derives new LQR control gains based on the measurement of rider weight at the initial calibration stage, and (2) the interaction compensation module that generates the full state command vector $\mathbf{s_{cy}^*} = [\theta_{EQy}, 0, 0, \dot\phi_{cy}]$ and compensation torque $\mathbf{\tau_{EQy}}$ for the low-level controller.}
  \label{block_diagram}
\end{figure}

The core of the PURE control system is the LQR-PI controller and a command speed generation module (Fig. \ref{block_diagram}a). The LQR-PI controller takes in the error signal ($\mathbf{s_{ey}} = \mathbf{s_{cy}} - \mathbf{s_y}$) to generate reference input torque ($\tau_{ry}$) for the model-reference PI controller. The command speed generation is the essence of the hands-free control scheme that outputs command SW angular speed ($\dot\phi_{cy}$) using physical human-robot interactions ($\mathbf{u_{py}}$) measured from Torso-dynamics Estimation System (TES) from the PURE hardware.

The proposed iHACS contains both offline (control gain personalization) and online (interaction compensation) modules that were integrated into the existing control system of PURE (Fig. \ref{block_diagram}b). The new controller of iHACS can be written as
\begin{equation}
        \tau_{ry} = \mathbf{k_{LQR}^*} (\mathbf{s_{cy}^*} - \mathbf{s_y}) + \tau_{EQy}
\end{equation}
where $\mathbf{k_{LQR}^*}$ are personalized control gains for the LQR controller, $\mathbf{s_y}$ is the feedback state vector. $\tau_{ry}$ is the reference torque applied to the ballbot, which is further utilized in the model-reference PI controller to compute the input torque $\tau_y$ for the control of PURE in the sagittal plane \cite{xiao2023design}.

\section{Human Subject Tests with Physical Hardware}

Human subject experiments were conducted to compare the speed-tracking performance of the rider-ballbot system using HACS and iHACS for simple SC tasks. Each controller was tested under two conditions: idle-keeping and speed-limiting tasks. Two able-bodied individuals (ABI) and two manual wheelchair users (mWCU) were recruited for this test (Table I). They all have experience with riding PURE with HACS in a previous study \cite{song2024driving}. The testing protocol of this study was approved by the institutional review board at the University of Illinois at Urbana-Champaign.

\begin{table}[]
    \centering
    \caption{Subject Demographics}
    \begin{tabular}{|c c c c c|}
        \hline
         Subject ID & Age &Gender & Height (m) & Weight (kg) \\
         \hline
         S04 (ABI) & 23 & F & 1.64 & 50 \\
         S07 (ABI) & 22 & M & 1.76 & 73 \\
         S12 (mWCU) & 21 & M & 1.60 & 79 \\
         S16 (mWCU) & 32 & F & 1.67 & 52 \\
         \hline
    \end{tabular}

    \label{tab3:sub_demo}
\end{table}

\subsection{Testing Protocol}

Each participant underwent a brief training session ($\sim$30 min) to reacquaint themselves with riding PURE, involving a braking task with both HACS and iHACS. In this task, participants were instructed to accelerate PURE to 1.4 m/s and brake to full stop upon receiving a visual cue. Initially, PURE was operated using HACS with the same tunable parameters identified in their previous experiment. Subsequently, participants rode PURE with iHACS applied to the sagittal plane control. A new set of sensitivity parameters was adjusted for the command speed generator in iHACS.

Following the training session, participants completed the idle-keeping and speed-limiting tasks using both HACS and iHACS for sagittal plane control. The order of testing HACS and iHACS was randomized for each participant. The participants were instructed to act "adversarially" against the objective of SC with PURE. That is, the rider and SC have different objectives. SC was set to idle or limit speed, while the rider was instructed to move PURE or to accelerate PURE to a high speed, respectively.

\begin{figure}[thpb]
  \centering
  \includegraphics[scale=0.5]{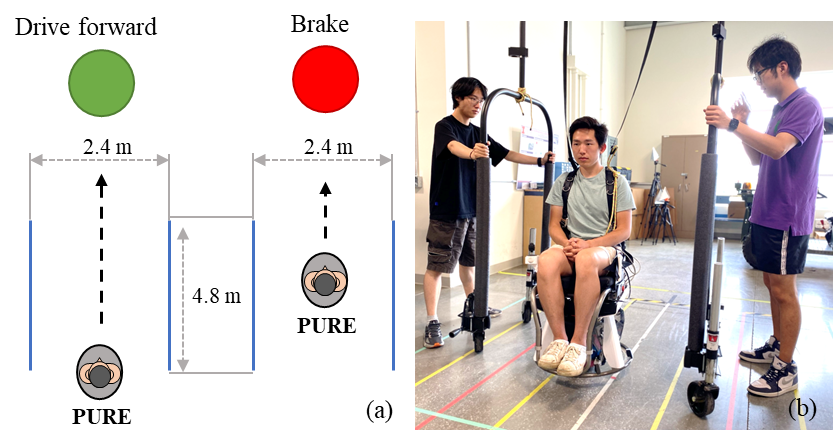}
  \caption{(a) Test course illustration of speed-limiting tasks. Green and red projections represented “go” and “stop” maneuvers, respectively. (b) Participant riding PURE with a mobile harness system. }
  \label{set_up}
\end{figure}

In the idle-keeping task, the command speed for the sagittal plane was set to zero regardless of the participant's torso leaning. Participants were instructed to lean their torso forward and backward as far as possible to induce movement in PURE until the ballbot started translating in any direction. This task was performed three times for each controller.

In the speed-limiting task, the SC limited the command speed to 0.5 m/s. Participants were asked to lean forward as far as possible to achieve their fastest translational speed (up to 1.4 m/s) while monitoring a projected circular image on the floor (Fig. \ref{set_up}a). They were instructed to execute a quick braking maneuver upon either: 1) crossing the finishing line, or 2) exceeding 1.4 m/s (indicated by red color). This task was also repeated three times for each controller.

During all activities, the participant wore a safety harness that was attached to a mobile gantry. Two researchers pushed the gantry to follow the participant (Fig. \ref{set_up}b)

\subsection{Data Collection and Processing}

To analyze the speed-tracking performance, drivetrain states ($\mathbf{s_y}$), rider’s toro lean angle ($\zeta_y$), and interaction forces and moments ($\mathbf{u_{py}}$) were recorded at 400 Hz. The maximum torso lean angle ($\zeta_y$), chassis tilt angle ($\theta_y$), and translation speed ($v_y = \dot\phi_{y}r_W$, $r_W$ is the spherical wheel radius) were obtained for each trial. Additionally, we calculated the root-mean-square error (RMSE) between the command and measured SW angular speed as follows: 
\begin{equation}
    RMSE = \frac{1}{t_f-t_0}\sqrt{\sum_{t_0}^{t_f} (\dot\phi_{cy}(t) - \dot\phi_y(t))^2}
    \label{eqn3:rmse}
\end{equation}
where $\dot\phi_{cy}(t)$ and $\dot\phi_y(t)$ are the command and measured SW angular speed, respectively. $t_0$ and $t_f$ are the start and end times of a selected trial.

\begin{figure}[thpb]
  \centering
  \includegraphics[scale=0.5]{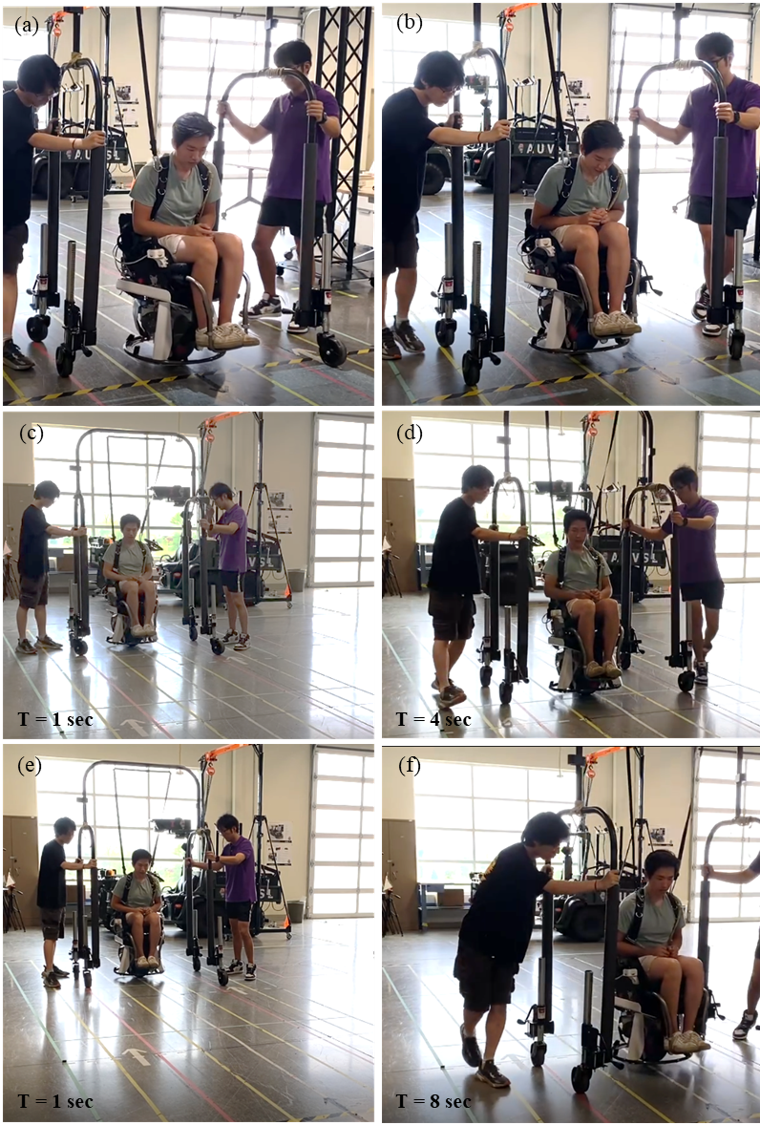}
  \caption{Idle-keeping and speed limiting tests of ABI participant. (a-b) Idle-keeping test with (a) HACS and (b) iHACS: maximum lean angle used immediately before PURE started to move; larger lean was required with iHACS. (c-d) Speed limiting test with HACS; top speed of 1.4 m/s was achieved after 3 s of effort. (e-f) Speed limiting test with iHACS; 1.4 m/s was still not achieved after 7 s. }
  \label{human_subject_test}
\end{figure}

We anticipated that iHACS would provide improved tracking of command speed to achieve the objective of the SC when the robot was subjected to adversarial torso leaning from participants. Consequently, the drivetrain would interactively tilt the chassis in the opposite direction of torso leaning to try to maintain the (small) commanded speed. In other words, participants would need to expend greater adversarial effort to overpower the robot under iHACS compared to HACS. Thus, we expected a larger maximum torso lean angle ($max|\zeta_y|$) and chassis tilt angle ($max|\theta_y|$), lower maximum translational speed ($max|v_y|$) and smaller RMSE when iHACS was used compared in both tasks. 

% would be better at resisting robot motion and require the rider to use larger maximum torso lean angle ($max|\zeta_y|$) and chassis tilt angle ($max|\theta_y|$), lower maximum translational speed ($max|\dot\phi_y|$), and smaller $RMSE$ between the command and measured translational speed compared to HACS in both tasks. 

\subsection{Results}

%Using iHACS required the rider to use a larger torso lean angle and the robot used a greater chassis tilt angle while limiting the robot to a lower maximum speed and a reduced RMSE in command speed tracking compared to HACS in both tasks (Table \ref{tab:results}). Screen shots from the video of a participant during the tests are also provided in Fig. \ref{human_subject_test} to visually demonstrate the effectiveness of iHACS.

\begin{figure}[thpb]
  \centering
  \includegraphics[scale=0.845]{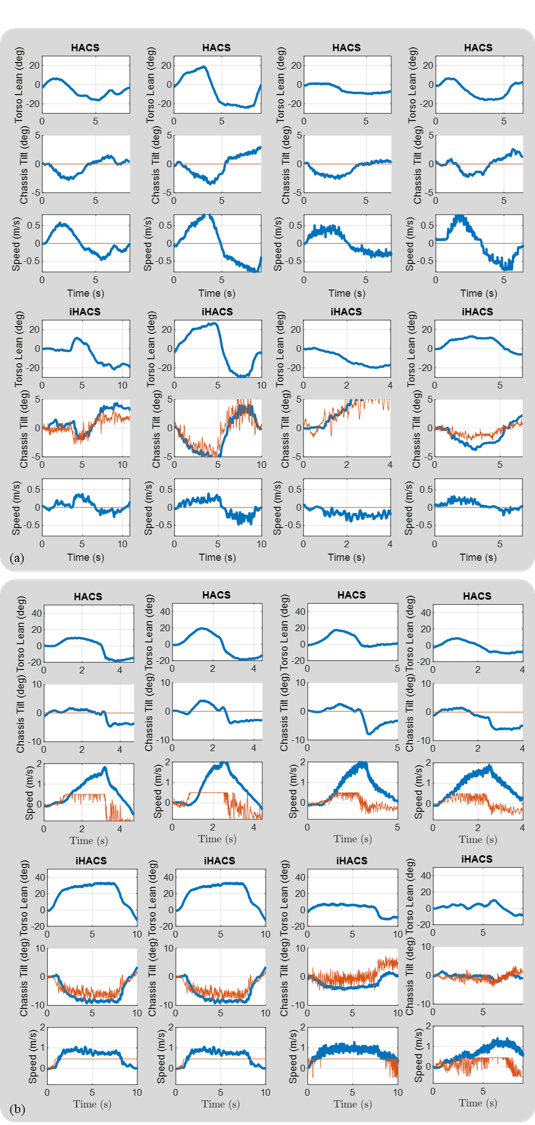}
  \caption{PURE system states and rider torso lean angle in (a)idle-keeping and (b) speed-limiting tasks for all four participants, one exemplary trial from each participant is presented.
  The top four plots are associated with HACS, and the bottom four plots are from iHACS trials. Blue lines: torso lean angle $\zeta_y$, chassis tilt angle $\theta_y$, and translational speed $v_y$. Orange lines: command chassis tilt angle $\theta_{EQy}$ and command translation speed $\dot\phi_{cy}$}
  \label{exp_trajectory}
\end{figure}

Generally, higher rider torso lean angle and chassis tilt angle, and lower translation speed and RMSE were observed when iHACS was used (Table \ref{tab:results}, Fig. \ref{human_subject_test}, Fig. \ref{exp_trajectory}). 
In the idling task with HACS, participants achieved an averaged maximum torso lean angle of 13.3$^\circ$ to generate a translational speed of 0.6 m/s. The robot chassis tilted backward for 2.6$^\circ$ but still failed to keep the user idling. When iHACS is utilized, the participants leaned almost twice as much (28.7$^\circ$) but only resulted in an average maximum speed of 0.3 m/s. Similar behaviors of the participants and PURE were observed in speed-limiting trials as well.

\begin{table}[]
    \caption{Results of Human Subject Test, Mean (SD)}
    \centering
    \begin{tabular}{|c|c c |c c |}
        \hline
        Tasks & \multicolumn{2}{c|} {Idle-Keeping} & \multicolumn{2}{c|} {Speed-Limiting}\\
        \hline
        Controller  & HACS & iHACS & HACS & iHACS  \\
        \hline
        $max|\zeta_y|$ (deg)  & 13.3 (6.2) & 28.7 (6.4) & 9.5 (5.8) & 18.5 (11.7)\\
        $max|\theta_y|$ (deg) & 2.6 (0.6) & 6.4(1.9) & 1.8 (0.6) & 3.9 (2.5) \\
        $max|v_y|$ (m/s) & 0.6 (0.2) & 0.3(0.0) & 1.9 (0.4) & 1.1 (0.2)\\
        $RMSE$ (E-3) & 7.0 (2.2)& 3.2 (1.2)& 21.2 (6.3)& 6.7 (2.1)\\
        \hline
    \end{tabular}
    
    \label{tab:results}
\end{table}

Since participants were instructed to act against the shared controller objectives, the increased torso lean angle observed with iHACS indicates greater resistance to disrupting the system’s adherence to objective of SC. The increased chassis tilt angle resulted from iHACS actively compensating for the applied interaction forces and torques by tilting the chassis in the opposite direction of torso leaning. The reduced translational speed and lower RMSE highlight the effectiveness of iHACS in achieving the SC goals of limiting translational speed while counteracting adversarial forces and torques. These improvements were less pronounced with HACS. These results demonstrated iHACS's enhanced capability for managing simple shared control tasks such as idle-keeping and speed-limiting.

\section{DISCUSSION}

This study introduced a controller that demonstrated higher performance by leveraging human-robot interaction between the rider and ballbot. 
Physical robot experiments were conducted with two able-bodied individuals and two manual wheelchair riders, with weight differences of up to 27 kg in each group. In the idle-keeping tasks, iHACS demonstrated minimal translational motion and low command speed tracking RMSE, even with significant torso lean angles.

Previous implementations of PURE with HACS only responded to command speeds and physical interaction forces and torques from the rider. With iHACS, PURE now effectively compensates for the rider's torso-leaning dynamics, providing improved command speed tracking, and enabling the system to take over control authority when needed. When faced with torso-leaning motions that conflicted with the shared controller’s commands, PURE adjusted its chassis tilt angle in the opposite direction of the torso lean (Fig. \ref{exp_trajectory}). Such behavior amplified the haptic feeling of riding a dynamically stable device, the rider can feel the intention of PURE through such bilateral physical human-robot interactions.
Additionally, iHACS proved effective in both idle-keeping and speed-limiting tasks, demonstrating its robustness and eliminating the need for task-specific controller adjustments.
This enhanced human-robot interaction established a foundation for effective shared control between the rider and PURE. The interactive hands-free admittance control scheme, iHACS, will allow the integration of advanced collision detection and avoidance systems developed for statically stable mobile devices \cite{herrera2018modeling,urdiales2011new,philips2007adaptive} into the dynamically stable PURE for future SC developments.

There are several limitations to this study. 
\subsubsection{Limited sample size}
The sample size was limited to four participants, preventing statistically significant conclusions about the performance of iHACS compared to HACS. 
\subsubsection{Speed tracking error} While iHACS improved performance, further reductions in command speed tracking errors are needed (Table \ref{tab:results}). Specifically, PURE still exceeded the 0.5 m/s limit during the speed-limiting task, suggesting that the equilibrium tilt angle and input torque may not be fully effective in dynamic scenarios. The LQR control gains were derived based on a linearized model around the upright equilibrium, which may not remain optimal as the system deviates from this equilibrium.
\subsubsection{Command signal noise}
The command state signals including $\dot\phi_{cy}$ and $\theta_{EQy}$ both have low signal-to-noise ratio, especially for manual wheelchair users who usually require high sensitivity parameters due to their limited torso range of motion. We did observe the system produced more chattering when iHACS was utilized. Load cells with analog outputs and signal amplifiers used in TES hardware contributed to a large portion of the noise.
\subsubsection{Simple SC tasks}
Only simple SC tasks were evaluated, where command speeds were relatively constant. Future work should explore iHACS performance in more complex SC scenarios involving static and moving obstacles.

Future directions include investigating time-varying LQR control gains and model predictive control schemes \cite{underactuated} to generate full-state trajectory to improve speed tracking performance in dynamic tasks like speed-limiting and assistive braking. 
Utilizing load cells with digital outputs could help to increase the signal-to-noise ratio and reduce the system chattering. Low-pass filtering could also be helpful to improve the behavior, but it should be carefully selected to not interfere with the coupled stability of the rider-ballbot system since it adds extra signal delay in the closed-loop control system.
Additionally, incorporating control of translation position could enhance position-keeping capabilities.

\section{CONCLUSION}

This study introduced and evaluated the interactive hands-free admittance control scheme (iHACS) for a riding ballbot. Building on the hands-free admittance control scheme (HACS) from a previous study, iHACS integrates an offline control-gain personalization module and an online interaction compensation module to address command speed tracking issues. This enhancement aimed to facilitate simple shared control tasks such as idle-keeping and speed-limiting.
%The physical robot experiments involved two able-bodied individuals and two manual wheelchair riders, with weight differences of up to 27 kg in each group. In the idle-keeping tasks, iHACS demonstrated minimal translational motion and low command speed tracking RMSE, even with significant torso lean angles. During the speed-limiting task, where the speed was capped at 0.5 m/s, participants could not exceed 0.75 m/s with iHACS, compared to speeds over 1.8 m/s with HACS. iHACS also achieved small command speed tracking RMSE despite large torso lean angles.
With iHACS, PURE transitioned from a system that passively responsed to physical interactions, to one that actively modulated system behavior based on physical interactions and speed commands from the shared controller. This capability enables the creation of an intelligent SC agent that enhances rider safety by managing intrinsic and extrinsic information to avoid collisions with static and dynamic obstacles, limit maximum speed, and reduce the proficiency required for hands-free control via torso motions.

\addtolength{\textheight}{-0cm}   % This command serves to balance the column lengths

%%%%%%%%%%%%%%%%%%%%%%%%%%%%%%%%%%%%%%%%%%%%%%%%%%%%%%%%%%%%%%%%%%%%%%%%%%%%%%%%

%%%%%%%%%%%%%%%%%%%%%%%%%%%%%%%%%%%%%%%%%%%%%%%%%%%%%%%%%%%%%%%%%%%%%%%%%%%%%%%%

%%%%%%%%%%%%%%%%%%%%%%%%%%%%%%%%%%%%%%%%%%%%%%%%%%%%%%%%%%%%%%%%%%%%%%%%%%%%%%%%
\section*{ACKNOWLEDGMENT}

The authors thank Doctor Jeannette Elliot, Professor Deana McDonagh, Doctor Patricia Malik, graduate students 
 Nadja Marin, and undergraduate students Chentai Yuan, Yixiang Guo for their help and support with concept development and human subject testing.

\bibliographystyle{IEEEtran}
\bibliography{IEEEabrv, reference}

\end{document}